\documentclass[10pt,twocolumn,letterpaper]{article}

\usepackage{cvpr}              

\usepackage[dvipsnames]{xcolor}

\definecolor{cvprblue}{rgb}{0.21,0.49,0.74}
\usepackage[pagebackref,breaklinks,colorlinks,citecolor=cvprblue]{hyperref}

\def\paperID{38} 
\def\confName{CVPR}
\def\confYear{2024}

\title{Pytorch-Wildlife: A Collaborative Deep Learning Framework for Conservation}

\author{Andres Hernandez* $^{1,2}$
\and
Zhongqi Miao* $^{1}$
\and
Luisa Vargas $^{1,2}$
\and
Sara Beery $^{3}$
\and
Rahul Dodhia $^{1}$
\and
Pablo Arbelaez $^{2}$
\and
Juan M. Lavista Ferres$^{1}$
\and
$^{1}$Microsoft AI for Good, Redmond.\\
$^{2}$Center for Research and Formation in Artificial Intelligence, Universidad de los Andes, Colombia.\\
$^{3}$Massachusetts Institute of Technology.\\
}
\begin{document}
\maketitle
\let\thefootnote\relax\footnote{*Equal contribution.}
\begin{abstract}
The alarming decline in global biodiversity, driven by various factors, underscores the urgent need for large-scale wildlife monitoring. In response, scientists have turned to automated deep learning methods for data processing in wildlife monitoring. However, applying these advanced methods in real-world scenarios is challenging due to their complexity and the need for specialized knowledge, primarily because of technical challenges and interdisciplinary barriers.

To address these challenges, we introduce Pytorch-Wildlife, an open-source deep learning platform built on PyTorch. It is designed for creating, modifying, and sharing powerful AI models. This platform emphasizes usability and accessibility, making it accessible to individuals with limited or no technical background. It also offers a modular codebase to simplify feature expansion and further development.

Pytorch-Wildlife offers an intuitive, user-friendly interface, accessible through local installation or Hugging Face, for animal detection and classification in images and videos. As two real-world applications, Pytorch-Wildlife has been utilized to train animal classification models for species recognition in the Amazon Rainforest and for invasive opossum recognition in the Galápagos Islands. The Opossum model achieves $98\%$ accuracy, and the Amazon model has $92\%$ recognition accuracy for $36$ animals in $90\%$ of the data. As Pytorch-Wildlife evolves, we aim to integrate more conservation tasks, addressing various environmental challenges. 

Pytorch-Wildlife is available at \url{https://github.com/microsoft/CameraTraps}.

\end{abstract}
\section{Introduction}
\label{sec:intro}

The role of wildlife in maintaining ecosystem balance is being jeopardized by human activities, underscoring an urgent need for large-scale biodiversity monitoring~\cite{declive1}. To surmount the logistic challenges of fieldwork and data collection, particularly in remote and megadiverse regions, automated data collection devices—such as camera traps~\cite{tobler2015spatiotemporal,mugerwa2023global, Zalan,coleman2023using}, autonomous recording units~\cite{Laiolo2010, MarquesThomasEtAl2013, SugaiSilvaEtAl2019}, and overhead cameras mounted on drones and satellites~\cite{kellenberger202121,miao2023challenges}—have been deployed. These tools have proven effective but generate vast datasets that require manual processing and annotation. This represents a critical bottleneck in data processing.

Deep learning technologies, such as Convolutional Neural Networks (CNNs)~\cite{alexnet, resnet}, have enhanced the ability to process large quantities of data with complex structures like images. Moreover, they show remarkable performance in animal detection and classification~\cite{megadetector,Miao,Zalan}. Yet, implementing these technologies in practical conservation efforts is challenging, especially to practitioners with limited engineering backgrounds. To effectively implement these approaches, we identify three key aspects that require consideration: accessibility, scalability, and transparency. First, accessibility refers to the ease with which these models can be utilized; they should have easy installation processes and be compatible across different operating systems. Also, the model should provide a user interface to facilitate its use by non-technical users. Secondly, scalability focuses on the framework's ability to add new features and to adapt to the user's needs, ensuring its applicability across various scenarios. Finally, transparency, on the other hand, advocates for a fully open-source framework, which allows users to understand the undergoing process at each step and enables the development of new projects based on this framework.


In recent years, various frameworks have been developed to address the challenges associated with wildlife imagery. For instance, Trapper~\cite{trapper}, Timelapse~\cite{timelapse} and Camelot~\cite{camelot} help users to easily store, annotate, manage, map and visualize camera trap imagery. Moreover, Zooniverse~\cite{zooniverse} promotes people-powered research to help scientists to annotate large volumes of data through the creation of projects. Furthermore, platforms such as Agouti~\cite{agouti}, Wildlife Insights~\cite{wildlifeinsights} and DeepLab Cut~\cite{deeplabcut} provide a complete set of tools that include automated wildlife recognitions through deep learning models. 


However, a significant challenge remains with the interoperability of these platforms with other open-source efforts and new deep learning models. The lack of access to the source code in frameworks like Wildlife Insights and Agouti, limits the ability to expand features and develop custom tools. This is particularly relevant in a diverse field like camera trap analysis, where there is a strong demand for flexible, customizable solutions. On the other hand some open-source tools like DeepLabCut have a very specialized and narrow focus (e.g., key-point detection), making it difficult to expand their area of focus and include a wide range of tools and models.

In this study, we introduce Pytorch-wildlife, an open-source deep-learning framework designed specifically for conservation that emphasizes on accessibility, scalability, and transparency. Pytorch-wildlife can be installed via pip on any operating system that supports Python, featuring a modular architecture that facilitates the rapid incorporation of new features, models, and datasets. This framework includes a comprehensive model zoo that provides various models for animal detection and classification alongside a user interface designed for non-technical users to interact with all of its features. Pytorch-Wildlife aims to provide a unified and versatile AI platform for the conservation community and to expand to a broader range of conservation tasks, including detection, classification, and segmentation, from various data sources such as camera traps, bioacoustics, and overhead imagery. Pytorch-wildlife is completely open source, and the code is available at \url{https://github.com/microsoft/CameraTraps}.\\

Our contributions can be summarized as follows:
\begin{itemize}
    \item We present a new open-source deep learning framework, Pytorch-Wildlife, tailored for conservation projects, and designed with a focus on accessibility, scalability, and transparency. The codebase of Pytorch-Wildife is modular, which significantly simplifies the process of integrating new features, models, and datasets.
    \item We showcase two real-world applications where we use Pytorch-Wildlife to detect and recognize animals for opossums in Galápagos Islands and for 36 animal genus in the Amazon Rainforest.
    \item We develop a user interface with Pytorch-Wildlife that allows for intuitive access to the framework’s capabilities.
\end{itemize}

\begin{figure*}[t]
\centering
  \includegraphics[width=0.75\linewidth]{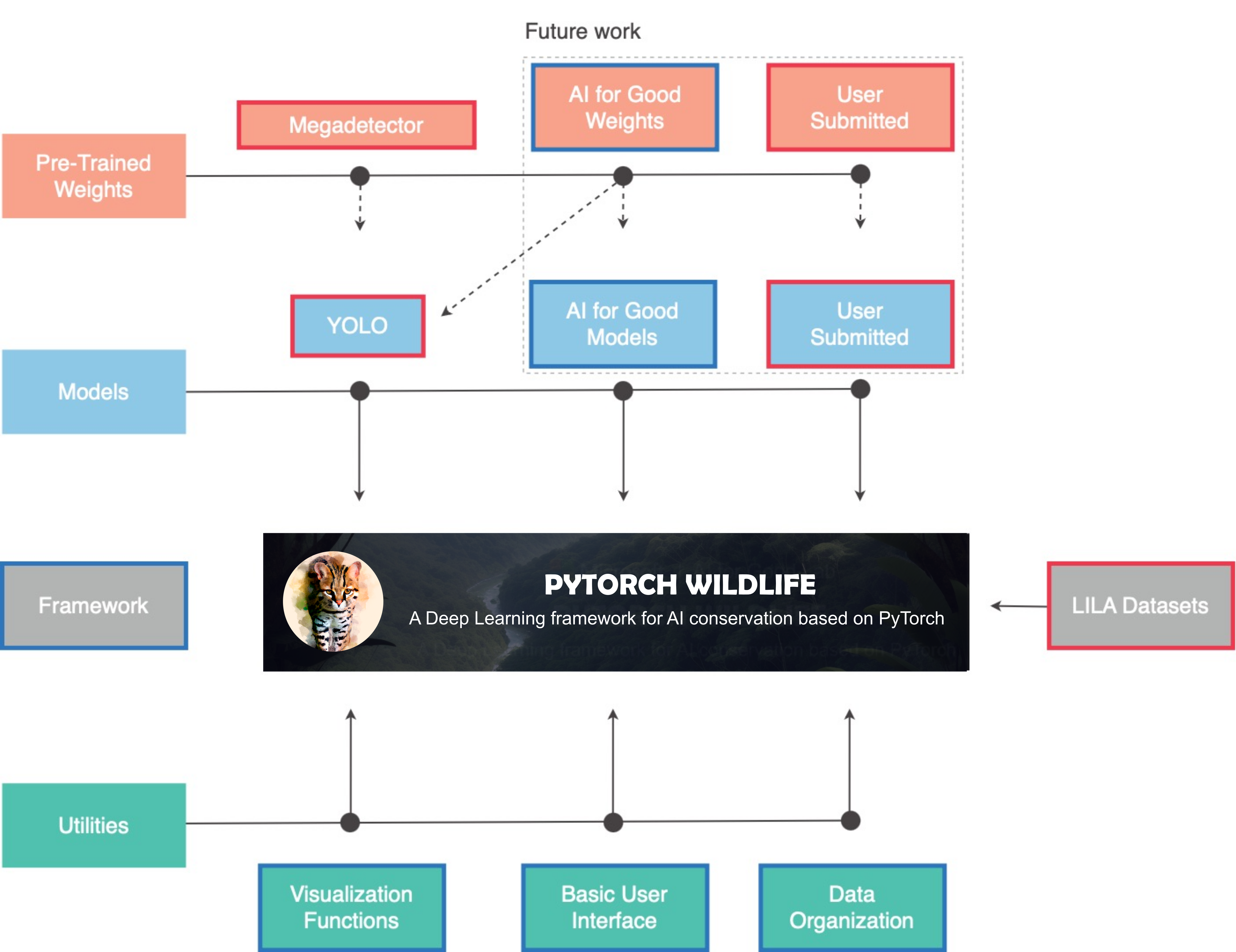}
  \captionof{figure}{Overview figure of the Pytorch-wildlife framework. First, it connects to the datasets available on LILA-BC for ease of training and validation; second, it offers detection and classification models with pretrained weights in a variety of datasets; finally, it comes with a user interface and a set of utility functions for visualization and adequate post-processing.}
  \label{fig:overview}
\quad
\end{figure*}
\section{Pytorch-Wildlife}
\label{sec:formatting}
\subsection*{Core components}

\begin{figure}[h]
\centering
  \includegraphics[width=0.99\linewidth]{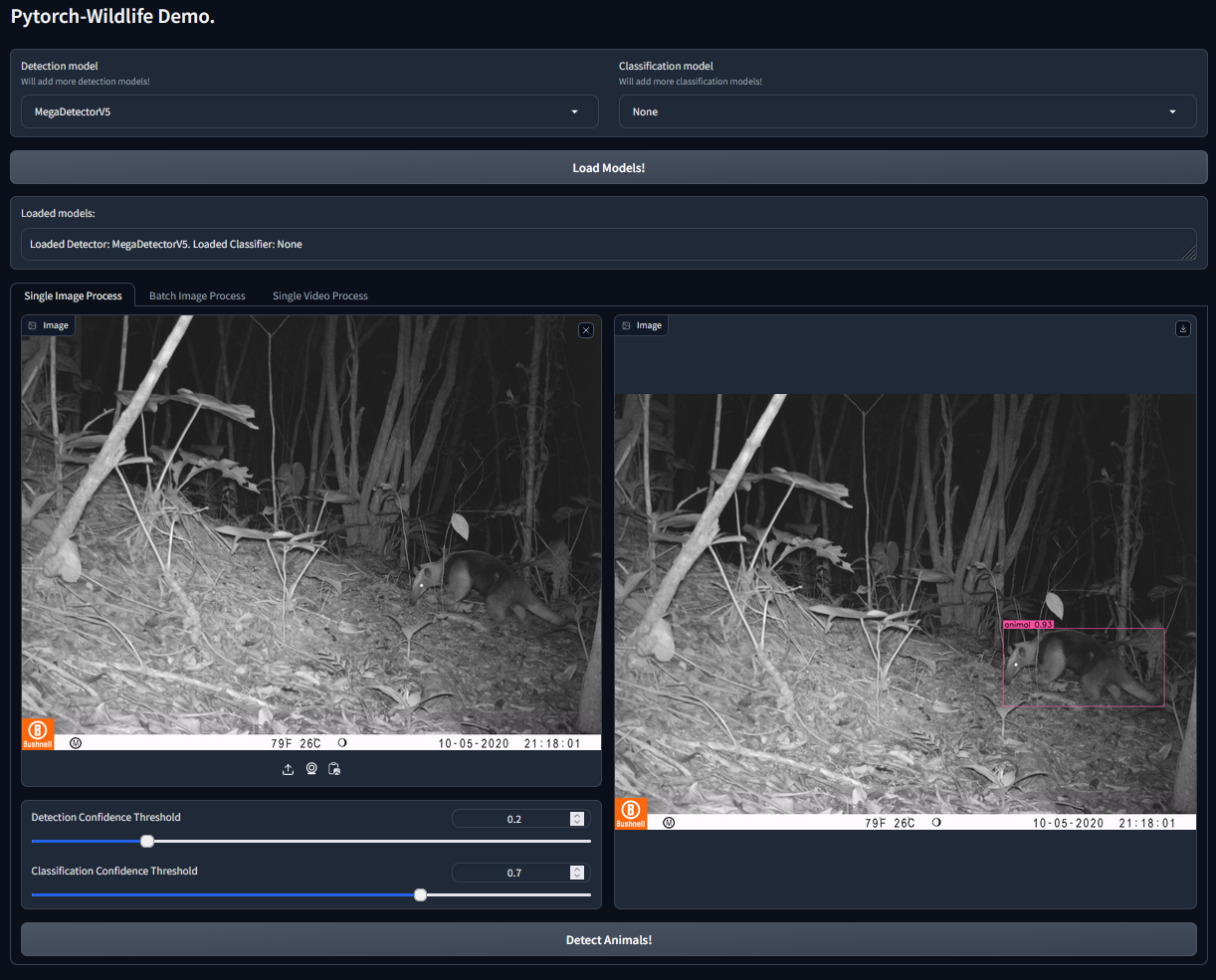}
  \captionof{figure}{Pytorch-wildlife user interface. It allows the user to load detection and classification models, as well as to perform single image detection, batch image detection, and single video detection with a confidence threshold for human validation.}
  \label{fig:ui}
\quad
\end{figure}

Figure~\ref{fig:overview} outlines the core components of Pytorch-Wildlife, which consists of three fundamental elements: 1) datasets, 2) a model zoo with model definitions and pre-trained weights, and 3) utility functions. First, to promote the creation of new AI models that are easy to replicate and test, Pytorch-wildlife connects users to the variety of open datasets available at the Labeled Information Library of Alexandria: Biology and Conservation (LILA-BC)~\cite{LILA}. Secondly, Pytorch-wildlife has a model zoo that currently has one detection model, MegaDetectorV5~\cite{megadetector}, and three animal recognition models. MegaDetectorV5 is the 5th version of MegaDetector, an animal detection model that is trained on 3 million animal images across different regions and ecosystems using YOLOv5 detection model architecture~\cite{yolov5}. The three animal recognition models in the current model zoo are trained on data collected from three different region: (i) the Amazon Rainforest, (ii) the Galápagos Islands, and (iii) the Serengeti National Park. Each recognition model is trained for a specific task: (i) genus classification, (ii) opossum classification, and (iii) species classification, respectively. For transparency and replicability, Pytorch-Wildlife provides both the model architectures and the pre-trained weights in the model zoo. In addition, Pytorch-Wildlife also provides a classification model fine-tuning module used to train these animal recognition models. Finally, the framework is rounded out with utility functions for data pre- and post-processing and data visualization. A user interface that ensures users, regardless of their technical skills, can access and test available features in Pytorch-Wildlife. Figure~\ref{fig:ui} shows the Pytorch-Wildlife user interface, here users can choose a detection and an optional classification model to perform single image detection, batch image detection and single video detection. Moreover, we enable direct control of detection and classification threshold to facilitate human-in-the-loop procedures. 

As stated previously, a complete framework should thrive on accessibility, scalability, and transparency. In the following subsection, we will review the Pytorch-wildlife features under each category.

\subsection*{Accessibility}

Accessibility is a key factor in integrating deep learning models into practical applications. Pytorch-wildlife has been developed with this in mind, ensuring it can be easily installed via pip and is compatible with any operating system that supports Python. Moreover, to aid users of diverse technical backgrounds, the installation process is accompanied by visual guides, Jupyter and Google Colab notebooks, and video tutorials. Additionally, the framework's models are optimized for local and low-end device inference to ensure that there is no need for an internet connection or a dedicated GPU. Also, for those who prefer not to use local resources, a version of Pytorch-wildlife is also hosted on HuggingFace (\url{https://huggingface.co/spaces/ai4g-biodiversity/pytorch-wildlife}), allowing for remote use of the framework's capabilities. 

\subsection*{Scalability}
The protocols and subjects of research in wildlife monitoring vary significantly depending on factors such as the type of study, the ecosystem, and the available tools. For this reason, the framework must be flexible to accommodate to distinct data structures. Moreover, it should offer versatile post-processing options to tailor outputs to the specific user requirements, as researchers may use supplementary software for human validation or to adhere to unique data structures. Pytorch-wildlife is designed with a modular codebase and includes utility functions for customized data splitting based on location, time, or season. Moreover, it features a default COCO output format that can be easily modified, enabling the saving of results as annotated images. Besides the default COCO format, Pytorch-Wildlife is also compatible with Timelapse (\url{https://saul.cpsc.ucalgary.ca/timelapse/}) and EcoAssist(\url{https://addaxdatascience.com/ecoassist/}) for people that prefer these data processing interfaces. Furthermore, Pytorch-Wildlife includes a classification fine-tuning module that allows users to train their own recognition models. Models produced by this fine-tuning module are compatible with the Pytorch-Wildlife framework and can be released through the Pytorch-Wildlife model zoo platform.

\subsection*{Transparency}
To ensure that a framework remains useful in a rapidly changing environment, it is important that its code is openly available to the public. This allows new features and models to be implemented easily and promotes community contribution. For this reason, Pytorch-wildlife adheres to these principles, with an open-source codebase hosted on GitHub. Finally, to assist users in leveraging the framework to its full potential, it provides a complete documentation set and technical support to ensure clarity and guidance for all levels of engagement.

\subsection*{Model Zoo}
As part of Pytorch-Wildlife's transparency initiative, we encourage contributors to send their own deep learning models to be readily available in our model zoo. Given the unique challenges of camera trap imagery, these detection and classification models have performance limitations tied to the training data. Variables such as the species in the dataset, the geographical location, the complexity of the background, the type of climate and the resolution of the camera, among others, can significantly affect the model's ability to predict a class.

To define the limitations of a model, we are establishing a leaderboard with standardized hidden test sets for animal detection and classification based on data from the LILA~\cite{LILA} dataset, this leaderboard shows the performance of the model on each hidden test, as well as a conclusion on the type of data that aligns best with the training data. Moreover, each model uploaded to the zoo through the community channel contains a 'community feedback' component, where verified users can rate and add comments on the limitations on the model. This leaderboard will help users to select the most appropriate model for their specific requirements, ensuring better applicability and reliability in their research.

\subsection*{MegaDetectorV6}

MegaDetectorV5~\cite{megadetector} is one of the golden standards in animal detection for camera trap imagery. However, its deep learning backbone is based on an older architecture, and newer detection models have achieved state-of-the-art performance on the detection task. Additionally, one of the main challenges in MegaDetectorV5 is that it is trained on the largest version of YOLOv5~\cite{yolov5}, which makes it incompatible with smaller devices and edge computing.

To address these limitations, we trained a new deep learning model using the YOLOv9-compact~\cite{yolov9} architecture on the MegaDetectorV5 training dataset. We call this model MegaDetectorV6-compact (MDv6-c). MDv6-c has only one-sixth of the parameters of MegaDetectorV5, making it more suitable for smaller devices. 

Table~\ref{tab:megadetector} shows the quantitative comparison between MegaDetectorV5 (MDv5) and MDv6-c. We calculate the precision and recall of these models using the MegaDetectorV5 validation dataset to ensure a fair comparison. MDv6-c achieves a recall rate of 0.85, which is 12 percentage points higher than MegaDetectorV5. Remarkably, this improvement is accomplished with a model that has only one-sixth of the parameters of MegaDetectorV5. The higher recall on this compact-sized model allows more animals to be detected and reducing false negatives. 


The compact nature of MegaDetectorV6-compact is particularly advantageous in resource-constrained environments and devices, where we can generate predictions faster and more efficient without compromising performance. In addition to improved recall, this model's efficiency paves the way for new features in Pytorch-Wildlife that are tailored for smaller devices. These enhancements will further extend the utility and versatility of our framework, supporting practitioners in diverse field conditions.\\

\begin{table}[h!]
  \centering
  \begin{tabular}{c c c c c}
    \toprule
    Model & Parameters & Precision & Recall & mAP\\
    \midrule
    MDv5 & 121M & 0.96 & 0.73 & 0.85\\
    MDv6-c & 22M & 0.92 & 0.85 & 0.84\\

    \bottomrule
  \end{tabular}
  \caption{Precision and Recall comparison between MegaDetectorV5 (MDv5) and MegaDetectorV6 trained with YOLOv9-compact (MDv6-c). MDv6-c increases 12 percentage points in the recall with one-sixth of the parameters. This makes it suitable for inference in smaller devices.}
  \label{tab:megadetector}
\end{table}

In the future, we are also providing a larger version of MegaDetectorV6 using YOLOv9-extra for optimal detection performance and also a transformer based MegaDetector using RT-DETR~\cite{rtdetr} to prepare MegaDetector for the future of transformer based models. 

MegaDetectorV6-compact's pretrained weights are available in Pytorch-Wildlife.





\section{Applications in real-life scenarios}
In this section, we demonstrate two real-world applications using Pytorch-Wildlife.

\subsection*{Background}

The Amazon Rainforest, spanning around $7$ million $km^2$ across Brazil, Perú, Colombia, Ecuador, Venezuela, Bolivia, Guyana, Suriname, and French Guyana, represents a vital component of the planet's biodiversity~\cite{Cepal}. The biome, however, faces significant threats from deforestation, primarily due to agriculture and livestock practices~\cite{Humboldt2}, leading to biodiversity loss and ecosystem degradation. Particularly, Colombia is one of the most affected countries, where its portion of the Amazon Rainforest experiences the highest deforestation rates within its territory~\cite{Humboldt2}. In the effort to survey biodiversity, camera traps have emerged as essential tools for monitoring fauna. These tools, however, pose a challenge in processing the vast amounts of data they generate, including distinguishing between relevant animal captures and false positives triggered by non-animal movements. Thus, there is a need for rapid and efficient analysis of this data to implement timely conservation strategies. In this project, we use Pytorch-Wildlife to separate animal and non-animal images and classify each animal into their genus taxonomic group.\\

Similarly, the Galápagos Islands represent another critical hotspot for biodiversity, where $97\%$ of terrestrial reptiles and mammals, $80\%$ of terrestrial birds, and more than $30\%$ of plant species are endemic of the archipelago~\cite{Galapagos1}. However, the fragile equilibrium of the Galápagos' biodiversity faces a critical threat from invasive species. The introduction of non-native species into a highly isolated ecosystem such as the Galápagos Islands can modify the population dynamics of local species and result in their extinction~\cite{Galapagos2}. As these invasive species can out-compete the endemic ones, continuous monitoring and management efforts are key to keeping the balance on the ecosystem. In this project, we use Pytorch-Wildlife for the detection of opossums (\textit{Didelphis spp.}) that get accidentally introduced during boat transportation. 

\subsection*{Datasets}

\begin{itemize}
    \item Amazon Rainforest camera trap dataset: The dataset for this project contains $41,904$ images across 36 labeled genera, and it is distributed into $33,569$ images in the training set and $8335$ images in the validation set.
    \item Galápagos Islands: The dataset for this project contains $491,471$ videos that are labeled as either "opossum" or "non-opossum". We split the dataset to contain $343,053$ videos in the training set and $148,418$ videos for the validation set. In order to process the videos in the pipeline, each video was split into image frames using a framerate of $30$ fps; if the video had a native fps lower than $30$, the native fps was used instead.
\end{itemize}

\subsection*{Experimental setup}
As an example of using Pytorch-Wildlife to detect and train project-dedicated animal classification models, we first perform detection inference using MegaDetectorV5 with Pytorch-Wildlife~\cite{megadetector} on all images in the datasets to filter out all the empty images and images with non-animal objects, such as human and vehicles. We then crop and resize the detected animal objects to $256\times256$ pixels and assign labels to each cropped image with image-level annotations. We use the classification fine-tuning module in Pytorch-Wildlife to train recognition models using these cropped images. The default model architecture is ResNet-50~\cite{resnet}, and the default classification training setting is $60$ training epochs with a batch size of $128$ using stochastic gradient descent~\cite{sgd} and a step learning rate scheduler with a step size of 20 epochs.

\subsection*{Quantitative results on the two datasets}

In our inference of the Amazon Rainforest dataset, we implement a 98\% confidence threshold as part of a human-in-the-loop procedure. We observe that the model predicts 90\% of the data with recognition confidence exceeding this threshold. Furthermore, within this high-confidence subset, the model achieves an average classification accuracy of 92\%. This means that, after filtering out empty images with MegaDetectorv5, only 10\% of the detected animal objects require human validation.

On the other hand, in the Galápagos dataset, we classify each video individually by applying majority voting on the prediction results for all of the frames within a video. The model achieves $98\%$ accuracy classification between opossums and other non-opossum animals. Both video processing and prediction majority voting are included in Pytorch-Wildlife, and the two models are being applied by our partners from both projects.

\section{Conclusions}
We present Pytorch-wildlife, an open-source deep learning framework designed for nature conservation. The framework is designed with a focus on accessibility, scalability, and transparency. As Pytorch-Wildlife expands, it aims to support an expansive array of conservation tasks and ultimately become a tool for advancing conservation efforts using AI across different environments and study areas

\section{Ethical considerations}

Sharing camera trap data with spatial metadata poses risks like exposing endangered species to poachers. To mitigate this, Pytorch-wildlife platform will generalize location information through a metadata filtering process. Additionally, to address privacy concerns, human images will be removed from datasets shared through the platform.

{
    \small
    \bibliographystyle{ieeenat_fullname}
    \bibliography{main}
}


\end{document}